\documentclass[10pt, a4paper]{article}
\usepackage{lrec2022} 
\usepackage{multibib}
\newcites{languageresource}{Language Resources}
\usepackage{graphicx}
\usepackage{tabularx}
\usepackage{enumitem}

\usepackage{soul}
\usepackage{titlesec}
\titleformat{\section}{\normalfont\large\bfseries\center}{\thesection.}{1em}{}
\titleformat{\subsection}{\normalfont\SmallTitleFont\bfseries\raggedright}{\thesubsection.}{1em}{}
\titleformat{\subsubsection}{\normalfont\normalsize\bfseries\raggedright}{\thesubsubsection.}{1em}{}
\renewcommand\thesection{\arabic{section}}
\renewcommand\thesubsection{\thesection.\arabic{subsection}}
\renewcommand\thesubsubsection{\thesubsection.\arabic{subsubsection}}

\usepackage{epstopdf}
\usepackage[utf8]{inputenc}

\usepackage[hyphens]{url}
\usepackage{hyperref}
\usepackage[hyphenbreaks]{breakurl}
\usepackage{xstring}

\usepackage{xcolor}
\usepackage{t1enc}
\usepackage{tikz-dependency}
\usepackage{qtree}
\usepackage{float}
\usepackage{tikz-qtree}

\usepackage{latexsym,amssymb}
\usepackage{caption}
\usepackage{subcaption}
\usepackage{tikz-qtree}
\usepackage{tikz}
\usepackage{color}
\usepackage{soul}
\usepackage{hyperref}
\usepackage{multirow}
\usepackage{hhline,colortbl}
\usepackage{siunitx}
\newcolumntype{s}[1]{S[table-format=#1]}
\DeclareSIUnit\k{k}
\DeclareSIUnit\M{M}
\DeclareSIUnit\G{G}

\pgfdeclarelayer{edgelayer}
\pgfdeclarelayer{nodelayer}
\pgfsetlayers{edgelayer,nodelayer,main}

\title{Dilated Convolutional Neural Networks \\ for Lightweight Diacritics Restoration}

\name{Bálint Csanády, András Lukács}
\address{Department of Computer Science, AI Research Group\\
         Institute of Mathematics, Eötvös Loránd University\\
         Pázmány Péter stny. 1/c, 1036 Budapest, Hungary\\
         csbalint@protonmail.ch, lukacs@cs.elte.hu}

\abstract{
Diacritics restoration has become a ubiquitous task in the Latin-alphabet-based English-dominated Internet language environment.
In this paper, we describe a small footprint 1D dilated convolution-based approach which operates on a character-level.
We find that solutions based on 1D dilated convolutional neural networks are competitive alternatives to models based on recursive neural networks or linguistic modeling for the task of diacritics restoration.
Our solution surpasses the performance of similarly sized models and is also competitive with larger models.
A special feature of our solution is that it even runs locally in a web browser.
We also provide a working example of this browser-based implementation. 
Our model is evaluated on different corpora, with emphasis on the Hungarian language.
We performed comparative measurements about the generalization power of the model in relation to three Hungarian corpora.
We also analyzed the errors to understand the limitation of corpus-based self-supervised training.
\\ \newline }

\begin{document}

\maketitleabstract

\section{Introduction}

Many languages, including most European languages, have alphabets where some of the characters are derived from base characters using \emph{diacritical marks}.
The goal of \emph{diacritics restoration} is to restore diacritical marks, given an input text which does not contain (or only partially contains) the proper diacritical marks.
Diacritics restoration is a practical task on the Internet, where the absence of diacritical marks can still be prominent.

Diacritics restoration is a useful preprocessing step for many NLP tasks, e.g. question answering \cite{abdelnasser2014bayan}.
On the other hand, diacritics restoration is an important tool for language revitalization \cite{galla2009indigenous},
thus contributing to linguistic diversity, the literacy of endangered languages,
and the maintenance of their digital presence \cite{kornai2013digital}.
This can be effectively supported by language-independent diacritics restoration tools.
Nevertheless, we consider only living languages where large corpora based on the Latin alphabet are available (omitting such exciting cases as Celtic languages or poetry marking).
Diacritical marks appear in certain Slavic languages (Czech, Slovak, Polish), some Finno-Ugric languages (Finnish, Hungarian, Latvian), Romanian, Turkish, and, most intensively, in Vietnamese.

Approaches to diacritics restoration have evolved from rule-based and statistical solutions to the application of machine learning models \cite{yarowsky1999comparison}.
The latter approach can be broken down into solutions using fixed or learned representations.
All solutions with learned representations seem to be based on neural networks connected to the models used in NLP, lately recurrent neural networks models \cite{hucko2018diacritics} being replaced by transformers \cite{pazmany,naplava2021diacritics}.
In such cases, models used for machine translation are often used to correct diacritical marks \cite{novak2015automatic}.
Another approach is to consider diacritics restoration as a sequence labeling problem where convolutional neural networks and recurrent neural networks such as BiLSTM-s \cite{naplava2018} can be applied.
We apply a fast language-independent method with small footprint for automatic diacritics restoration using a neural architecture based on 1D convolutions, the so called Acausal Temporal Convolutional Networks (A-TCN).
Models based on A-TCN have comparable performance to BiLSTM-s \cite{A-TCN}, which is also confirmed by our present research.

Our experiments are focused on the Hungarian language. In Hungarian the characters which can receive diacritical marks are exactly the vowels (e.g. u $\mapsto \{$u,ú,ü,ű$\}$). 
For Hungarian, the current state of the art is reported by \cite{pazmany} and is achieved by neural machine translation.   
Our main contribution is a lightweight model, which can even be run locally in web browsers, allowing client-side inference.
We compared our model with Hunaccent \cite{hunaccent}; both models have a similar size of around 10MB.
Our approach outperforms Hunaccent by a large margin, and performs comparable to the recurrent model in \newcite{naplava2018}.
Moreover our method is also language-agnostic.

\section{Methods}

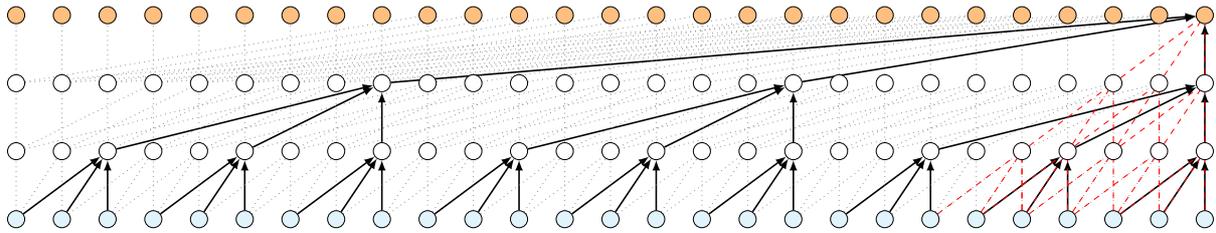
\begin{figure*}[ht!]
    \centering
    \resizebox{\textwidth}{!}{
\begin{tikzpicture}[scale=.4]
	\begin{pgfonlayer}{nodelayer}
		\node [circle, draw=black, fill={cyan!10}, minimum size=3mm, inner sep=0pt] (0) at (0, 0) {};
		\node [circle, draw=black, fill={cyan!10}, minimum size=3mm, inner sep=0pt] (1) at (2, 0) {};
		\node [circle, draw=black, fill={cyan!10}, minimum size=3mm, inner sep=0pt] (2) at (4, 0) {};
		\node [circle, draw=black, fill={cyan!10}, minimum size=3mm, inner sep=0pt] (3) at (6, 0) {};
		\node [circle, draw=black, fill={cyan!10}, minimum size=3mm, inner sep=0pt] (4) at (8, 0) {};
		\node [circle, draw=black, fill={cyan!10}, minimum size=3mm, inner sep=0pt] (5) at (10, 0) {};
		\node [circle, draw=black, fill={cyan!10}, minimum size=3mm, inner sep=0pt] (6) at (12, 0) {};
		\node [circle, draw=black, fill={cyan!10}, minimum size=3mm, inner sep=0pt] (7) at (14, 0) {};
		\node [circle, draw=black, fill={cyan!10}, minimum size=3mm, inner sep=0pt] (8) at (16, 0) {};
		\node [circle, draw=black, fill={cyan!10}, minimum size=3mm, inner sep=0pt] (9) at (18, 0) {};
		\node [circle, draw=black, fill={cyan!10}, minimum size=3mm, inner sep=0pt] (10) at (20, 0) {};
		\node [circle, draw=black, fill={cyan!10}, minimum size=3mm, inner sep=0pt] (11) at (22, 0) {};
		\node [circle, draw=black, fill={cyan!10}, minimum size=3mm, inner sep=0pt] (12) at (24, 0) {};
		\node [circle, draw=black, fill={cyan!10}, minimum size=3mm, inner sep=0pt] (13) at (26, 0) {};
		\node [circle, draw=black, fill={cyan!10}, minimum size=3mm, inner sep=0pt] (14) at (28, 0) {};
		\node [circle, draw=black, fill={cyan!10}, minimum size=3mm, inner sep=0pt] (15) at (30, 0) {};
		\node [circle, draw=black, fill={cyan!10}, minimum size=3mm, inner sep=0pt] (16) at (32, 0) {};
		\node [circle, draw=black, fill={cyan!10}, minimum size=3mm, inner sep=0pt] (17) at (34, 0) {};
		\node [circle, draw=black, fill={cyan!10}, minimum size=3mm, inner sep=0pt] (18) at (36, 0) {};
		\node [circle, draw=black, fill={cyan!10}, minimum size=3mm, inner sep=0pt] (19) at (38, 0) {};
		\node [circle, draw=black, fill={cyan!10}, minimum size=3mm, inner sep=0pt] (20) at (40, 0) {};
		\node [circle, draw=black, fill={cyan!10}, minimum size=3mm, inner sep=0pt] (21) at (42, 0) {};
		\node [circle, draw=black, fill={cyan!10}, minimum size=3mm, inner sep=0pt] (22) at (44, 0) {};
		\node [circle, draw=black, fill={cyan!10}, minimum size=3mm, inner sep=0pt] (23) at (46, 0) {};
		\node [circle, draw=black, fill={cyan!10}, minimum size=3mm, inner sep=0pt] (24) at (48, 0) {};
		\node [circle, draw=black, fill={cyan!10}, minimum size=3mm, inner sep=0pt] (25) at (50, 0) {};
		\node [circle, draw=black, fill={cyan!10}, minimum size=3mm, inner sep=0pt] (26) at (52, 0) {};
		\node [circle, draw=black, fill=white, minimum size=3mm, inner sep=0pt] (27) at (0, 3) {};
		\node [circle, draw=black, fill=white, minimum size=3mm, inner sep=0pt] (28) at (2, 3) {};
		\node [circle, draw=black, fill=white, minimum size=3mm, inner sep=0pt] (29) at (4, 3) {};
		\node [circle, draw=black, fill=white, minimum size=3mm, inner sep=0pt] (30) at (6, 3) {};
		\node [circle, draw=black, fill=white, minimum size=3mm, inner sep=0pt] (31) at (8, 3) {};
		\node [circle, draw=black, fill=white, minimum size=3mm, inner sep=0pt] (32) at (10, 3) {};
		\node [circle, draw=black, fill=white, minimum size=3mm, inner sep=0pt] (33) at (12, 3) {};
		\node [circle, draw=black, fill=white, minimum size=3mm, inner sep=0pt] (34) at (14, 3) {};
		\node [circle, draw=black, fill=white, minimum size=3mm, inner sep=0pt] (35) at (16, 3) {};
		\node [circle, draw=black, fill=white, minimum size=3mm, inner sep=0pt] (36) at (18, 3) {};
		\node [circle, draw=black, fill=white, minimum size=3mm, inner sep=0pt] (37) at (20, 3) {};
		\node [circle, draw=black, fill=white, minimum size=3mm, inner sep=0pt] (38) at (22, 3) {};
		\node [circle, draw=black, fill=white, minimum size=3mm, inner sep=0pt] (39) at (24, 3) {};
		\node [circle, draw=black, fill=white, minimum size=3mm, inner sep=0pt] (40) at (26, 3) {};
		\node [circle, draw=black, fill=white, minimum size=3mm, inner sep=0pt] (41) at (28, 3) {};
		\node [circle, draw=black, fill=white, minimum size=3mm, inner sep=0pt] (42) at (30, 3) {};
		\node [circle, draw=black, fill=white, minimum size=3mm, inner sep=0pt] (43) at (32, 3) {};
		\node [circle, draw=black, fill=white, minimum size=3mm, inner sep=0pt] (44) at (34, 3) {};
		\node [circle, draw=black, fill=white, minimum size=3mm, inner sep=0pt] (45) at (36, 3) {};
		\node [circle, draw=black, fill=white, minimum size=3mm, inner sep=0pt] (46) at (38, 3) {};
		\node [circle, draw=black, fill=white, minimum size=3mm, inner sep=0pt] (47) at (40, 3) {};
		\node [circle, draw=black, fill=white, minimum size=3mm, inner sep=0pt] (48) at (42, 3) {};
		\node [circle, draw=black, fill=white, minimum size=3mm, inner sep=0pt] (49) at (44, 3) {};
		\node [circle, draw=black, fill=white, minimum size=3mm, inner sep=0pt] (50) at (46, 3) {};
		\node [circle, draw=black, fill=white, minimum size=3mm, inner sep=0pt] (51) at (48, 3) {};
		\node [circle, draw=black, fill=white, minimum size=3mm, inner sep=0pt] (52) at (50, 3) {};
		\node [circle, draw=black, fill=white, minimum size=3mm, inner sep=0pt] (53) at (52, 3) {};
		\node [circle, draw=black, fill=white, minimum size=3mm, inner sep=0pt] (54) at (0, 6) {};
		\node [circle, draw=black, fill=white, minimum size=3mm, inner sep=0pt] (55) at (2, 6) {};
		\node [circle, draw=black, fill=white, minimum size=3mm, inner sep=0pt] (56) at (4, 6) {};
		\node [circle, draw=black, fill=white, minimum size=3mm, inner sep=0pt] (57) at (6, 6) {};
		\node [circle, draw=black, fill=white, minimum size=3mm, inner sep=0pt] (58) at (8, 6) {};
		\node [circle, draw=black, fill=white, minimum size=3mm, inner sep=0pt] (59) at (10, 6) {};
		\node [circle, draw=black, fill=white, minimum size=3mm, inner sep=0pt] (60) at (12, 6) {};
		\node [circle, draw=black, fill=white, minimum size=3mm, inner sep=0pt] (61) at (14, 6) {}; 

		\node [circle, draw=black, fill=white, minimum size=3mm, inner sep=0pt] (62) at (16, 6) {};
		\node [circle, draw=black, fill=white, minimum size=3mm, inner sep=0pt] (63) at (18, 6) {};
		\node [circle, draw=black, fill=white, minimum size=3mm, inner sep=0pt] (64) at (20, 6) {};
		\node [circle, draw=black, fill=white, minimum size=3mm, inner sep=0pt] (65) at (22, 6) {};
		\node [circle, draw=black, fill=white, minimum size=3mm, inner sep=0pt] (66) at (24, 6) {};
		\node [circle, draw=black, fill=white, minimum size=3mm, inner sep=0pt] (67) at (26, 6) {};
		\node [circle, draw=black, fill=white, minimum size=3mm, inner sep=0pt] (68) at (28, 6) {};
		\node [circle, draw=black, fill=white, minimum size=3mm, inner sep=0pt] (69) at (30, 6) {};
		\node [circle, draw=black, fill=white, minimum size=3mm, inner sep=0pt] (70) at (32, 6) {};
		\node [circle, draw=black, fill=white, minimum size=3mm, inner sep=0pt] (71) at (34, 6) {};
		\node [circle, draw=black, fill=white, minimum size=3mm, inner sep=0pt] (72) at (36, 6) {};
		\node [circle, draw=black, fill=white, minimum size=3mm, inner sep=0pt] (73) at (38, 6) {};
		\node [circle, draw=black, fill=white, minimum size=3mm, inner sep=0pt] (74) at (40, 6) {};
		\node [circle, draw=black, fill=white, minimum size=3mm, inner sep=0pt] (75) at (42, 6) {};
		\node [circle, draw=black, fill=white, minimum size=3mm, inner sep=0pt] (76) at (44, 6) {};
		\node [circle, draw=black, fill=white, minimum size=3mm, inner sep=0pt] (77) at (46, 6) {};
		\node [circle, draw=black, fill=white, minimum size=3mm, inner sep=0pt] (78) at (48, 6) {};
		\node [circle, draw=black, fill=white, minimum size=3mm, inner sep=0pt] (79) at (50, 6) {};
		\node [circle, draw=black, fill=white, minimum size=3mm, inner sep=0pt] (80) at (52, 6) {};
		\node [circle, draw=black, fill={orange!50}, minimum size=3mm, inner sep=0pt] (81) at (0, 9) {};
		\node [circle, draw=black, fill={orange!50}, minimum size=3mm, inner sep=0pt] (82) at (2, 9) {};
		\node [circle, draw=black, fill={orange!50}, minimum size=3mm, inner sep=0pt] (83) at (4, 9) {};
		\node [circle, draw=black, fill={orange!50}, minimum size=3mm, inner sep=0pt] (84) at (6, 9) {};
		\node [circle, draw=black, fill={orange!50}, minimum size=3mm, inner sep=0pt] (85) at (8, 9) {};
		\node [circle, draw=black, fill={orange!50}, minimum size=3mm, inner sep=0pt] (86) at (10, 9) {};
		\node [circle, draw=black, fill={orange!50}, minimum size=3mm, inner sep=0pt] (87) at (12, 9) {};
		\node [circle, draw=black, fill={orange!50}, minimum size=3mm, inner sep=0pt] (88) at (14, 9) {};
		\node [circle, draw=black, fill={orange!50}, minimum size=3mm, inner sep=0pt] (89) at (16, 9) {};
		\node [circle, draw=black, fill={orange!50}, minimum size=3mm, inner sep=0pt] (90) at (18, 9) {};
		\node [circle, draw=black, fill={orange!50}, minimum size=3mm, inner sep=0pt] (91) at (20, 9) {};
		\node [circle, draw=black, fill={orange!50}, minimum size=3mm, inner sep=0pt] (92) at (22, 9) {};
		\node [circle, draw=black, fill={orange!50}, minimum size=3mm, inner sep=0pt] (93) at (24, 9) {};
		\node [circle, draw=black, fill={orange!50}, minimum size=3mm, inner sep=0pt] (94) at (26, 9) {};
		\node [circle, draw=black, fill={orange!50}, minimum size=3mm, inner sep=0pt] (95) at (28, 9) {};
		\node [circle, draw=black, fill={orange!50}, minimum size=3mm, inner sep=0pt] (96) at (30, 9) {};
		\node [circle, draw=black, fill={orange!50}, minimum size=3mm, inner sep=0pt] (97) at (32, 9) {};
		\node [circle, draw=black, fill={orange!50}, minimum size=3mm, inner sep=0pt] (98) at (34, 9) {};
		\node [circle, draw=black, fill={orange!50}, minimum size=3mm, inner sep=0pt] (99) at (36, 9) {};
		\node [circle, draw=black, fill={orange!50}, minimum size=3mm, inner sep=0pt] (100) at (38, 9) {};
		\node [circle, draw=black, fill={orange!50}, minimum size=3mm, inner sep=0pt] (101) at (40, 9) {};
		\node [circle, draw=black, fill={orange!50}, minimum size=3mm, inner sep=0pt] (102) at (42, 9) {};
		\node [circle, draw=black, fill={orange!50}, minimum size=3mm, inner sep=0pt] (103) at (44, 9) {};
		\node [circle, draw=black, fill={orange!50}, minimum size=3mm, inner sep=0pt] (104) at (46, 9) {};
		\node [circle, draw=black, fill={orange!50}, minimum size=3mm, inner sep=0pt] (105) at (48, 9) {};
		\node [circle, draw=black, fill={orange!50}, minimum size=3mm, inner sep=0pt] (106) at (50, 9) {};
		\node [circle, draw=black, fill={orange!50}, minimum size=3mm, inner sep=0pt] (107) at (52, 9) {};
	\end{pgfonlayer}
	\begin{pgfonlayer}{edgelayer}
		\draw [draw, thick, -latex] (0) to (29);
		\draw [draw, thick, -latex] (1) to (29);
		\draw [draw, thick, -latex] (2) to (29);
		\draw [draw, thick, -latex] (3) to (32);
		\draw [draw, thick, -latex] (4) to (32);
		\draw [draw, thick, -latex] (5) to (32);
		\draw [draw, thick, -latex] (6) to (35);
		\draw [draw, thick, -latex] (7) to (35);
		\draw [draw, thick, -latex] (8) to (35);
		\draw [draw, thick, -latex] (9) to (38);
		\draw [draw, thick, -latex] (10) to (38);
		\draw [draw, thick, -latex] (11) to (38);
		\draw [draw, thick, -latex] (12) to (41);
		\draw [draw, thick, -latex] (13) to (41);
		\draw [draw, thick, -latex] (14) to (41);
		\draw [draw, thick, -latex] (15) to (44);
		\draw [draw, thick, -latex] (16) to (44);
		\draw [draw, thick, -latex] (17) to (44);
		\draw [draw, thick, -latex] (18) to (47);
		\draw [draw, thick, -latex] (19) to (47);
		\draw [draw, thick, -latex] (20) to (47);
		\draw [draw, thick, -latex] (21) to (50);
		\draw [draw, thick, -latex] (22) to (50);
		\draw [draw, thick, -latex] (23) to (50);
		\draw [draw, thick, -latex] (24) to (53);
		\draw [draw, thick, -latex] (25) to (53);
		\draw [draw, thick, -latex] (26) to (53);
		\draw [draw, thick, -latex] (29) to (62);
		\draw [draw, thick, -latex] (32) to (62);
		\draw [draw, thick, -latex] (35) to (62);
		\draw [draw, thick, -latex] (38) to (71);
		\draw [draw, thick, -latex] (41) to (71);
		\draw [draw, thick, -latex] (44) to (71);
		\draw [draw, thick, -latex] (47) to (80);
		\draw [draw, thick, -latex] (50) to (80);
		\draw [draw, thick, -latex] (53) to (80);
		\draw [draw, thick, -latex] (62) to (107);
		\draw [draw, thick, -latex] (71) to (107);
		\draw [draw, thick, -latex] (80) to (107);
		\draw [draw, dotted, -, gray] (0) to (27);
		\draw [draw, dotted, -, gray] (0) to (28);
		\draw [draw, dotted, -, gray] (1) to (28);
		\draw [draw, dotted, -, gray] (2) to (30);
		\draw [draw, dotted, -, gray] (2) to (31);
		\draw [draw, dotted, -, gray] (3) to (30);
		\draw [draw, dotted, -, gray] (3) to (31);
		\draw [draw, dotted, -, gray] (4) to (31);
		\draw [draw, dotted, -, gray] (4) to (33);
		\draw [draw, dotted, -, gray] (5) to (33);
		\draw [draw, dotted, -, gray] (6) to (33);
		\draw [draw, dotted, -, gray] (6) to (34);
		\draw [draw, dotted, -, gray] (7) to (34);
		\draw [draw, dotted, -, gray] (8) to (36);
		\draw [draw, dotted, -, gray] (8) to (37);
		\draw [draw, dotted, -, gray] (9) to (36);
		\draw [draw, dotted, -, gray] (9) to (37);
		\draw [draw, dotted, -, gray] (10) to (37);
		\draw [draw, dotted, -, gray] (10) to (39);
		\draw [draw, dotted, -, gray] (11) to (39);
		\draw [draw, dotted, -, gray] (12) to (39);
		\draw [draw, dotted, -, gray] (12) to (40);
		\draw [draw, dotted, -, gray] (13) to (40);
		\draw [draw, dotted, -, gray] (11) to (40);
		\draw [draw, dotted, -, gray] (7) to (36);
		\draw [draw, dotted, -, gray] (5) to (34);
		\draw [draw, dotted, -, gray] (1) to (30);
		\draw [draw, dotted, -, gray] (13) to (42);
		\draw [draw, dotted, -, gray] (14) to (42);
		\draw [draw, dotted, -, gray] (15) to (42);
		\draw [draw, dotted, -, gray] (15) to (43);
		\draw [draw, dotted, -, gray] (16) to (43);
		\draw [draw, dotted, -, gray] (14) to (43);
		\draw [draw, dotted, -, gray] (16) to (45);
		\draw [draw, dotted, -, gray] (17) to (45);
		\draw [draw, dotted, -, gray] (18) to (45);
		\draw [draw, dotted, -, gray] (18) to (46);
		\draw [draw, dotted, -, gray] (17) to (46);
		\draw [draw, dotted, -, gray] (19) to (46);
		\draw [draw, dotted, -, gray] (19) to (48);
		\draw [draw, dotted, -, gray] (20) to (48);
		\draw [draw, dotted, -, gray] (21) to (48);
		\draw [draw, dotted, -, gray] (20) to (49);
		\draw [draw, dotted, -, gray] (21) to (49);
		\draw [draw, dotted, -, gray] (22) to (49);
		\draw [draw, dotted, -, gray] (22) to (51);
		\draw [draw, dotted, -, gray] (23) to (51);
		\draw [draw, dotted, -, gray] (23) to (52);
		\draw [draw, dotted, -, gray] (24) to (51);
		\draw [draw, dotted, -, gray] (24) to (52);
		\draw [draw, dotted, -, gray] (25) to (52);
		\draw [draw, dotted, -, gray] (52) to (79);
		\draw [draw, dotted, -, gray] (49) to (79);
		\draw [draw, dotted, -, gray] (46) to (79);
		\draw [draw, dotted, -, gray] (51) to (78);
		\draw [draw, dotted, -, gray] (48) to (78);
		\draw [draw, dotted, -, gray] (45) to (78);
		\draw [draw, dotted, -, gray] (50) to (77);
		\draw [draw, dotted, -, gray] (47) to (77);
		\draw [draw, dotted, -, gray] (44) to (77);
		\draw [draw, dotted, -, gray] (49) to (76);
		\draw [draw, dotted, -, gray] (46) to (76);
		\draw [draw, dotted, -, gray] (43) to (76);
		\draw [draw, dotted, -, gray] (48) to (75);
		\draw [draw, dotted, -, gray] (45) to (75);
		\draw [draw, dotted, -, gray] (42) to (75);
		\draw [draw, dotted, -, gray] (47) to (74);
		\draw [draw, dotted, -, gray] (44) to (74);
		\draw [draw, dotted, -, gray] (41) to (74);
		\draw [draw, dotted, -, gray] (46) to (73);
		\draw [draw, dotted, -, gray] (43) to (73);
		\draw [draw, dotted, -, gray] (40) to (73);
		\draw [draw, dotted, -, gray] (45) to (72);
		\draw [draw, dotted, -, gray] (42) to (72);
		\draw [draw, dotted, -, gray] (39) to (72);
		\draw [draw, dotted, -, gray] (43) to (70);
		\draw [draw, dotted, -, gray] (40) to (70);
		\draw [draw, dotted, -, gray] (37) to (70);
		\draw [draw, dotted, -, gray] (42) to (69);
		\draw [draw, dotted, -, gray] (39) to (69);
		\draw [draw, dotted, -, gray] (36) to (69);
		\draw [draw, dotted, -, gray] (41) to (68);
		\draw [draw, dotted, -, gray] (38) to (68);
		\draw [draw, dotted, -, gray] (35) to (68);
		\draw [draw, dotted, -, gray] (40) to (67);
		\draw [draw, dotted, -, gray] (37) to (67);
		\draw [draw, dotted, -, gray] (34) to (67);
		\draw [draw, dotted, -, gray] (39) to (66);
		\draw [draw, dotted, -, gray] (36) to (66);
		\draw [draw, dotted, -, gray] (33) to (66);
		\draw [draw, dotted, -, gray] (38) to (65);
		\draw [draw, dotted, -, gray] (35) to (65);
		\draw [draw, dotted, -, gray] (32) to (65);
		\draw [draw, dotted, -, gray] (37) to (64);
		\draw [draw, dotted, -, gray] (36) to (63);
		\draw [draw, dotted, -, gray] (34) to (61);
		\draw [draw, dotted, -, gray] (33) to (60);
		\draw [draw, dotted, -, gray] (32) to (59);
		\draw [draw, dotted, -, gray] (31) to (58);
		\draw [draw, dotted, -, gray] (30) to (57);
		\draw [draw, dotted, -, gray] (29) to (56);
		\draw [draw, dotted, -, gray] (28) to (55);
		\draw [draw, dotted, -, gray] (27) to (54);
		\draw [draw, dotted, -, gray] (34) to (64);
		\draw [draw, dotted, -, gray] (33) to (63);
		\draw [draw, dotted, -, gray] (31) to (61);
		\draw [draw, dotted, -, gray] (30) to (60);
		\draw [draw, dotted, -, gray] (29) to (59);
		\draw [draw, dotted, -, gray] (28) to (58);
		\draw [draw, dotted, -, gray] (27) to (57);
		\draw [draw, dotted, -, gray] (31) to (64);
		\draw [draw, dotted, -, gray] (30) to (63);
		\draw [draw, dotted, -, gray] (28) to (61);
		\draw [draw, dotted, -, gray] (27) to (60);
		\draw [draw, dotted, -, gray] (61) to (106);
		\draw [draw, dotted, -, gray] (60) to (105);
		\draw [draw, dotted, -, gray] (59) to (104);
		\draw [draw, dotted, -, gray] (58) to (103);
		\draw [draw, dotted, -, gray] (57) to (102);
		\draw [draw, dotted, -, gray] (56) to (101);
		\draw [draw, dotted, -, gray] (55) to (100);
		\draw [draw, dotted, -, gray] (54) to (99);
		\draw [draw, dotted, -, gray] (70) to (106);
		\draw [draw, dotted, -, gray] (69) to (105);
		\draw [draw, dotted, -, gray] (68) to (104);
		\draw [draw, dotted, -, gray] (67) to (103);
		\draw [draw, dotted, -, gray] (66) to (102);
		\draw [draw, dotted, -, gray] (65) to (101);
		\draw [draw, dotted, -, gray] (64) to (100);
		\draw [draw, dotted, -, gray] (63) to (99);
		\draw [draw, dotted, -, gray] (62) to (98);
		\draw [draw, dotted, -, gray] (61) to (97);
		\draw [draw, dotted, -, gray] (60) to (96);
		\draw [draw, dotted, -, gray] (59) to (95);
		\draw [draw, dotted, -, gray] (58) to (94);
		\draw [draw, dotted, -, gray] (57) to (93);
		\draw [draw, dotted, -, gray] (56) to (92);
		\draw [draw, dotted, -, gray] (55) to (91);
		\draw [draw, dotted, -, gray] (54) to (90);
		\draw [draw, dotted, -, gray] (79) to (106);
		\draw [draw, dotted, -, gray] (78) to (105);
		\draw [draw, dotted, -, gray] (77) to (104);
		\draw [draw, dotted, -, gray] (76) to (103);
		\draw [draw, dotted, -, gray] (75) to (102);
		\draw [draw, dotted, -, gray] (74) to (101);
		\draw [draw, dotted, -, gray] (73) to (100);
		\draw [draw, dotted, -, gray] (72) to (99);
		\draw [draw, dotted, -, gray] (71) to (98);
		\draw [draw, dotted, -, gray] (70) to (97);
		\draw [draw, dotted, -, gray] (69) to (96);
		\draw [draw, dotted, -, gray] (68) to (95);
		\draw [draw, dotted, -, gray] (67) to (94);
		\draw [draw, dotted, -, gray] (66) to (93);
		\draw [draw, dotted, -, gray] (65) to (92);
		\draw [draw, dotted, -, gray] (64) to (91);
		\draw [draw, dotted, -, gray] (63) to (90);
		\draw [draw, dotted, -, gray] (62) to (89);
		\draw [draw, dotted, -, gray] (61) to (88);
		\draw [draw, dotted, -, gray] (60) to (87);
		\draw [draw, dotted, -, gray] (59) to (86);
		\draw [draw, dotted, -, gray] (58) to (85);
		\draw [draw, dotted, -, gray] (57) to (84);
		\draw [draw, dotted, -, gray] (56) to (83);
		\draw [draw, dotted, -, gray] (55) to (82);
		\draw [draw, dotted, -, gray] (54) to (81);
		\draw [draw, dashed, -, red] (78) to (107);
		\draw [draw, dashed, -, red] (79) to (107);
		\draw [draw, dashed, -, red] (80) to (107);
		\draw [draw, dashed, -, red] (51) to (80);
		\draw [draw, dashed, -, red] (52) to (80);
		\draw [draw, dashed, -, red] (53) to (80);
		\draw [draw, dashed, -, red] (50) to (79);
		\draw [draw, dashed, -, red] (51) to (79);
		\draw [draw, dashed, -, red] (52) to (79);
		\draw [draw, dashed, -, red] (49) to (78);
		\draw [draw, dashed, -, red] (50) to (78);
		\draw [draw, dashed, -, red] (51) to (78);
		\draw [draw, dashed, -, red] (20) to (49);
		\draw [draw, dashed, -, red] (21) to (49);
		\draw [draw, dashed, -, red] (22) to (49);
		\draw [draw, dashed, -, red] (22) to (49);
		\draw [draw, dashed, -, red] (21) to (50);
		\draw [draw, dashed, -, red] (22) to (50);
		\draw [draw, dashed, -, red] (23) to (50);
		\draw [draw, dashed, -, red] (22) to (51);
		\draw [draw, dashed, -, red] (23) to (51);
		\draw [draw, dashed, -, red] (24) to (51);
		\draw [draw, dashed, -, red] (23) to (52);
		\draw [draw, dashed, -, red] (24) to (52);
		\draw [draw, dashed, -, red] (25) to (52);
		\draw [draw, dashed, -, red] (24) to (53);
		\draw [draw, dashed, -, red] (25) to (53);
		\draw [draw, dashed, -, red] (26) to (53);
	\end{pgfonlayer}
\end{tikzpicture}
}
\caption{TCN architecture (kernel size: 3, dilation factors: 1,3,9). Red dashed: without dilation. }
\label{fig:TCN}
\end{figure*}

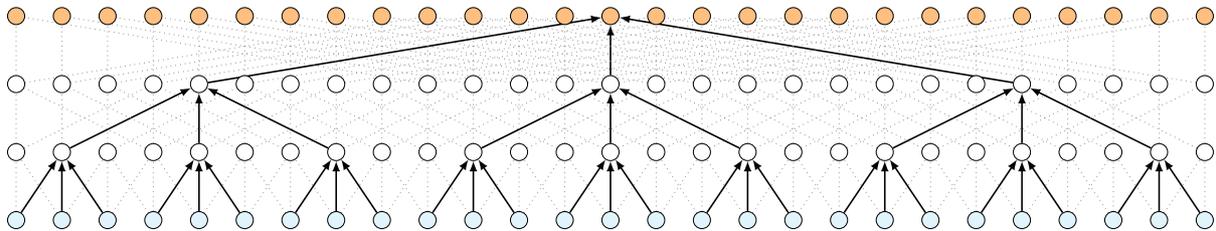
\begin{figure*}[ht!]
    \centering
    \resizebox{\textwidth}{!}{
\begin{tikzpicture}[scale=.4]
	\begin{pgfonlayer}{nodelayer}
		\node [circle, draw=black, fill={cyan!10}, minimum size=3mm, inner sep=0pt] (0) at (0, 0) {};
		\node [circle, draw=black, fill={cyan!10}, minimum size=3mm, inner sep=0pt] (1) at (2, 0) {};
		\node [circle, draw=black, fill={cyan!10}, minimum size=3mm, inner sep=0pt] (2) at (4, 0) {};
		\node [circle, draw=black, fill={cyan!10}, minimum size=3mm, inner sep=0pt] (3) at (6, 0) {};
		\node [circle, draw=black, fill={cyan!10}, minimum size=3mm, inner sep=0pt] (4) at (8, 0) {};
		\node [circle, draw=black, fill={cyan!10}, minimum size=3mm, inner sep=0pt] (5) at (10, 0) {};
		\node [circle, draw=black, fill={cyan!10}, minimum size=3mm, inner sep=0pt] (6) at (12, 0) {};
		\node [circle, draw=black, fill={cyan!10}, minimum size=3mm, inner sep=0pt] (7) at (14, 0) {};
		\node [circle, draw=black, fill={cyan!10}, minimum size=3mm, inner sep=0pt] (8) at (16, 0) {};
		\node [circle, draw=black, fill={cyan!10}, minimum size=3mm, inner sep=0pt] (9) at (18, 0) {};
		\node [circle, draw=black, fill={cyan!10}, minimum size=3mm, inner sep=0pt] (10) at (20, 0) {};
		\node [circle, draw=black, fill={cyan!10}, minimum size=3mm, inner sep=0pt] (11) at (22, 0) {};
		\node [circle, draw=black, fill={cyan!10}, minimum size=3mm, inner sep=0pt] (12) at (24, 0) {};
		\node [circle, draw=black, fill={cyan!10}, minimum size=3mm, inner sep=0pt] (13) at (26, 0) {};
		\node [circle, draw=black, fill={cyan!10}, minimum size=3mm, inner sep=0pt] (14) at (28, 0) {};
		\node [circle, draw=black, fill={cyan!10}, minimum size=3mm, inner sep=0pt] (15) at (30, 0) {};
		\node [circle, draw=black, fill={cyan!10}, minimum size=3mm, inner sep=0pt] (16) at (32, 0) {};
		\node [circle, draw=black, fill={cyan!10}, minimum size=3mm, inner sep=0pt] (17) at (34, 0) {};
		\node [circle, draw=black, fill={cyan!10}, minimum size=3mm, inner sep=0pt] (18) at (36, 0) {};
		\node [circle, draw=black, fill={cyan!10}, minimum size=3mm, inner sep=0pt] (19) at (38, 0) {};
		\node [circle, draw=black, fill={cyan!10}, minimum size=3mm, inner sep=0pt] (20) at (40, 0) {};
		\node [circle, draw=black, fill={cyan!10}, minimum size=3mm, inner sep=0pt] (21) at (42, 0) {};
		\node [circle, draw=black, fill={cyan!10}, minimum size=3mm, inner sep=0pt] (22) at (44, 0) {};
		\node [circle, draw=black, fill={cyan!10}, minimum size=3mm, inner sep=0pt] (23) at (46, 0) {};
		\node [circle, draw=black, fill={cyan!10}, minimum size=3mm, inner sep=0pt] (24) at (48, 0) {};
		\node [circle, draw=black, fill={cyan!10}, minimum size=3mm, inner sep=0pt] (25) at (50, 0) {};
		\node [circle, draw=black, fill={cyan!10}, minimum size=3mm, inner sep=0pt] (26) at (52, 0) {};
		\node [circle, draw=black, fill=white, minimum size=3mm, inner sep=0pt] (28) at (0, 3) {};
		\node [circle, draw=black, fill=white, minimum size=3mm, inner sep=0pt] (29) at (2, 3) {};
		\node [circle, draw=black, fill=white, minimum size=3mm, inner sep=0pt] (30) at (4, 3) {};
		\node [circle, draw=black, fill=white, minimum size=3mm, inner sep=0pt] (31) at (6, 3) {};
		\node [circle, draw=black, fill=white, minimum size=3mm, inner sep=0pt] (32) at (8, 3) {};
		\node [circle, draw=black, fill=white, minimum size=3mm, inner sep=0pt] (33) at (10, 3) {};
		\node [circle, draw=black, fill=white, minimum size=3mm, inner sep=0pt] (34) at (12, 3) {};
		\node [circle, draw=black, fill=white, minimum size=3mm, inner sep=0pt] (35) at (14, 3) {};
		\node [circle, draw=black, fill=white, minimum size=3mm, inner sep=0pt] (36) at (16, 3) {};
		\node [circle, draw=black, fill=white, minimum size=3mm, inner sep=0pt] (37) at (18, 3) {};
		\node [circle, draw=black, fill=white, minimum size=3mm, inner sep=0pt] (38) at (20, 3) {};
		\node [circle, draw=black, fill=white, minimum size=3mm, inner sep=0pt] (39) at (22, 3) {};
		\node [circle, draw=black, fill=white, minimum size=3mm, inner sep=0pt] (40) at (24, 3) {};
		\node [circle, draw=black, fill=white, minimum size=3mm, inner sep=0pt] (41) at (26, 3) {};
		\node [circle, draw=black, fill=white, minimum size=3mm, inner sep=0pt] (42) at (28, 3) {};
		\node [circle, draw=black, fill=white, minimum size=3mm, inner sep=0pt] (43) at (30, 3) {};
		\node [circle, draw=black, fill=white, minimum size=3mm, inner sep=0pt] (44) at (32, 3) {};
		\node [circle, draw=black, fill=white, minimum size=3mm, inner sep=0pt] (45) at (34, 3) {};
		\node [circle, draw=black, fill=white, minimum size=3mm, inner sep=0pt] (46) at (36, 3) {};
		\node [circle, draw=black, fill=white, minimum size=3mm, inner sep=0pt] (47) at (38, 3) {};
		\node [circle, draw=black, fill=white, minimum size=3mm, inner sep=0pt] (48) at (40, 3) {};
		\node [circle, draw=black, fill=white, minimum size=3mm, inner sep=0pt] (49) at (42, 3) {};
		\node [circle, draw=black, fill=white, minimum size=3mm, inner sep=0pt] (50) at (44, 3) {};
		\node [circle, draw=black, fill=white, minimum size=3mm, inner sep=0pt] (51) at (46, 3) {};
		\node [circle, draw=black, fill=white, minimum size=3mm, inner sep=0pt] (52) at (48, 3) {};
		\node [circle, draw=black, fill=white, minimum size=3mm, inner sep=0pt] (53) at (50, 3) {};
		\node [circle, draw=black, fill=white, minimum size=3mm, inner sep=0pt] (58) at (0, 6) {};
		\node [circle, draw=black, fill=white, minimum size=3mm, inner sep=0pt] (59) at (2, 6) {};
		\node [circle, draw=black, fill=white, minimum size=3mm, inner sep=0pt] (60) at (4, 6) {};
		\node [circle, draw=black, fill=white, minimum size=3mm, inner sep=0pt] (61) at (6, 6) {};
		\node [circle, draw=black, fill=white, minimum size=3mm, inner sep=0pt] (62) at (8, 6) {};
		\node [circle, draw=black, fill=white, minimum size=3mm, inner sep=0pt] (63) at (10, 6) {};
		\node [circle, draw=black, fill=white, minimum size=3mm, inner sep=0pt] (64) at (12, 6) {};
		\node [circle, draw=black, fill=white, minimum size=3mm, inner sep=0pt] (65) at (14, 6) {};
		\node [circle, draw=black, fill=white, minimum size=3mm, inner sep=0pt] (66) at (16, 6) {};
		\node [circle, draw=black, fill=white, minimum size=3mm, inner sep=0pt] (67) at (18, 6) {};
		\node [circle, draw=black, fill=white, minimum size=3mm, inner sep=0pt] (68) at (20, 6) {};
		\node [circle, draw=black, fill=white, minimum size=3mm, inner sep=0pt] (69) at (22, 6) {};
		\node [circle, draw=black, fill=white, minimum size=3mm, inner sep=0pt] (70) at (24, 6) {};
		\node [circle, draw=black, fill=white, minimum size=3mm, inner sep=0pt] (71) at (26, 6) {};
		\node [circle, draw=black, fill=white, minimum size=3mm, inner sep=0pt] (72) at (28, 6) {};
		\node [circle, draw=black, fill=white, minimum size=3mm, inner sep=0pt] (73) at (30, 6) {};
		\node [circle, draw=black, fill=white, minimum size=3mm, inner sep=0pt] (74) at (32, 6) {};
		\node [circle, draw=black, fill=white, minimum size=3mm, inner sep=0pt] (75) at (34, 6) {};
		\node [circle, draw=black, fill=white, minimum size=3mm, inner sep=0pt] (76) at (36, 6) {};
		\node [circle, draw=black, fill=white, minimum size=3mm, inner sep=0pt] (77) at (38, 6) {};
		\node [circle, draw=black, fill=white, minimum size=3mm, inner sep=0pt] (78) at (40, 6) {};
		\node [circle, draw=black, fill=white, minimum size=3mm, inner sep=0pt] (79) at (42, 6) {};
		\node [circle, draw=black, fill=white, minimum size=3mm, inner sep=0pt] (80) at (44, 6) {};
		\node [circle, draw=black, fill={orange!50}, minimum size=3mm, inner sep=0pt] (81) at (28, 9) {};
		\node [circle, draw=black, fill={orange!50}, minimum size=3mm, inner sep=0pt] (82) at (30, 9) {};
		\node [circle, draw=black, fill={orange!50}, minimum size=3mm, inner sep=0pt] (83) at (32, 9) {};
		\node [circle, draw=black, fill={orange!50}, minimum size=3mm, inner sep=0pt] (84) at (34, 9) {};
		\node [circle, draw=black, fill={orange!50}, minimum size=3mm, inner sep=0pt] (85) at (36, 9) {};
		\node [circle, draw=black, fill={orange!50}, minimum size=3mm, inner sep=0pt] (86) at (38, 9) {};
		\node [circle, draw=black, fill={orange!50}, minimum size=3mm, inner sep=0pt] (87) at (40, 9) {};
		\node [circle, draw=black, fill={orange!50}, minimum size=3mm, inner sep=0pt] (88) at (42, 9) {};
		\node [circle, draw=black, fill={orange!50}, minimum size=3mm, inner sep=0pt] (89) at (44, 9) {};
		\node [circle, draw=black, fill={orange!50}, minimum size=3mm, inner sep=0pt] (90) at (46, 9) {};
		\node [circle, draw=black, fill={orange!50}, minimum size=3mm, inner sep=0pt] (91) at (48, 9) {};
		\node [circle, draw=black, fill={orange!50}, minimum size=3mm, inner sep=0pt] (92) at (50, 9) {};
		\node [circle, draw=black, fill={orange!50}, minimum size=3mm, inner sep=0pt] (93) at (52, 9) {};
		\node [circle, draw=black, fill={orange!50}, minimum size=3mm, inner sep=0pt] (94) at (0, 9) {};
		\node [circle, draw=black, fill={orange!50}, minimum size=3mm, inner sep=0pt] (95) at (2, 9) {};
		\node [circle, draw=black, fill={orange!50}, minimum size=3mm, inner sep=0pt] (96) at (4, 9) {};
		\node [circle, draw=black, fill={orange!50}, minimum size=3mm, inner sep=0pt] (97) at (6, 9) {};
		\node [circle, draw=black, fill={orange!50}, minimum size=3mm, inner sep=0pt] (98) at (8, 9) {};
		\node [circle, draw=black, fill={orange!50}, minimum size=3mm, inner sep=0pt] (99) at (10, 9) {};
		\node [circle, draw=black, fill={orange!50}, minimum size=3mm, inner sep=0pt] (100) at (12, 9) {};
		\node [circle, draw=black, fill={orange!50}, minimum size=3mm, inner sep=0pt] (101) at (14, 9) {};
		\node [circle, draw=black, fill={orange!50}, minimum size=3mm, inner sep=0pt] (102) at (16, 9) {};
		\node [circle, draw=black, fill={orange!50}, minimum size=3mm, inner sep=0pt] (103) at (18, 9) {};
		\node [circle, draw=black, fill={orange!50}, minimum size=3mm, inner sep=0pt] (104) at (20, 9) {};
		\node [circle, draw=black, fill={orange!50}, minimum size=3mm, inner sep=0pt] (105) at (22, 9) {};
		\node [circle, draw=black, fill={orange!50}, minimum size=3mm, inner sep=0pt] (106) at (24, 9) {};
		\node [circle, draw=black, fill={orange!50}, minimum size=3mm, inner sep=0pt] (107) at (26, 9) {};
		\node [circle, draw=black, fill=white, minimum size=3mm, inner sep=0pt] (108) at (52, 3) {};
		\node [circle, draw=black, fill=white, minimum size=3mm, inner sep=0pt] (109) at (46, 6) {};
		\node [circle, draw=black, fill=white, minimum size=3mm, inner sep=0pt] (110) at (48, 6) {};
		\node [circle, draw=black, fill=white, minimum size=3mm, inner sep=0pt] (111) at (50, 6) {};
		\node [circle, draw=black, fill=white, minimum size=3mm, inner sep=0pt] (112) at (52, 6) {};
	\end{pgfonlayer}
	\begin{pgfonlayer}{edgelayer}
		\draw [draw, thick, -latex] (0) to (29);
		\draw [draw, thick, -latex] (1) to (29);
		\draw [draw, thick, -latex] (2) to (29);
		\draw [draw, thick, -latex] (3) to (32);
		\draw [draw, thick, -latex] (4) to (32);
		\draw [draw, thick, -latex] (5) to (32);
		\draw [draw, thick, -latex] (6) to (35);
		\draw [draw, thick, -latex] (7) to (35);
		\draw [draw, thick, -latex] (8) to (35);
		\draw [draw, thick, -latex] (9) to (38);
		\draw [draw, thick, -latex] (10) to (38);
		\draw [draw, thick, -latex] (11) to (38);
		\draw [draw, thick, -latex] (12) to (41);
		\draw [draw, thick, -latex] (13) to (41);
		\draw [draw, thick, -latex] (14) to (41);
		\draw [draw, thick, -latex] (15) to (44);
		\draw [draw, thick, -latex] (16) to (44);
		\draw [draw, thick, -latex] (17) to (44);
		\draw [draw, thick, -latex] (18) to (47);
		\draw [draw, thick, -latex] (19) to (47);
		\draw [draw, thick, -latex] (20) to (47);
		\draw [draw, thick, -latex] (21) to (50);
		\draw [draw, thick, -latex] (22) to (50);
		\draw [draw, thick, -latex] (23) to (50);
		\draw [draw, thick, -latex] (24) to (53);
		\draw [draw, thick, -latex] (25) to (53);
		\draw [draw, thick, -latex] (26) to (53);
		\draw [draw, thick, -latex] (29) to (62);
		\draw [draw, thick, -latex] (32) to (62);
		\draw [draw, thick, -latex] (35) to (62);
		\draw [draw, thick, -latex] (38) to (71);
		\draw [draw, thick, -latex] (41) to (71);
		\draw [draw, thick, -latex] (44) to (71);
		\draw [draw, thick, -latex] (47) to (80);
		\draw [draw, thick, -latex] (50) to (80);
		\draw [draw, thick, -latex] (53) to (80);
		\draw [draw, thick, -latex] (62) to (107);
		\draw [draw, thick, -latex] (71) to (107);
		\draw [draw, thick, -latex] (80) to (107);
		\draw [draw, dotted, -, gray] (0) to (28);
		\draw [draw, dotted, -, gray] (1) to (28);
		\draw [draw, dotted, -, gray] (2) to (30);
		\draw [draw, dotted, -, gray] (2) to (31);
		\draw [draw, dotted, -, gray] (3) to (30);
		\draw [draw, dotted, -, gray] (3) to (31);
		\draw [draw, dotted, -, gray] (4) to (31);
		\draw [draw, dotted, -, gray] (4) to (33);
		\draw [draw, dotted, -, gray] (5) to (33);
		\draw [draw, dotted, -, gray] (6) to (33);
		\draw [draw, dotted, -, gray] (6) to (34);
		\draw [draw, dotted, -, gray] (7) to (34);
		\draw [draw, dotted, -, gray] (8) to (36);
		\draw [draw, dotted, -, gray] (8) to (37);
		\draw [draw, dotted, -, gray] (9) to (36);
		\draw [draw, dotted, -, gray] (9) to (37);
		\draw [draw, dotted, -, gray] (10) to (37);
		\draw [draw, dotted, -, gray] (10) to (39);
		\draw [draw, dotted, -, gray] (11) to (39);
		\draw [draw, dotted, -, gray] (12) to (39);
		\draw [draw, dotted, -, gray] (12) to (40);
		\draw [draw, dotted, -, gray] (13) to (40);
		\draw [draw, dotted, -, gray] (11) to (40);
		\draw [draw, dotted, -, gray] (7) to (36);
		\draw [draw, dotted, -, gray] (5) to (34);
		\draw [draw, dotted, -, gray] (1) to (30);
		\draw [draw, dotted, -, gray] (13) to (42);
		\draw [draw, dotted, -, gray] (14) to (42);
		\draw [draw, dotted, -, gray] (15) to (42);
		\draw [draw, dotted, -, gray] (15) to (43);
		\draw [draw, dotted, -, gray] (16) to (43);
		\draw [draw, dotted, -, gray] (14) to (43);
		\draw [draw, dotted, -, gray] (16) to (45);
		\draw [draw, dotted, -, gray] (17) to (45);
		\draw [draw, dotted, -, gray] (18) to (45);
		\draw [draw, dotted, -, gray] (18) to (46);
		\draw [draw, dotted, -, gray] (17) to (46);
		\draw [draw, dotted, -, gray] (19) to (46);
		\draw [draw, dotted, -, gray] (19) to (48);
		\draw [draw, dotted, -, gray] (20) to (48);
		\draw [draw, dotted, -, gray] (21) to (48);
		\draw [draw, dotted, -, gray] (20) to (49);
		\draw [draw, dotted, -, gray] (21) to (49);
		\draw [draw, dotted, -, gray] (22) to (49);
		\draw [draw, dotted, -, gray] (22) to (51);
		\draw [draw, dotted, -, gray] (23) to (51);
		\draw [draw, dotted, -, gray] (23) to (52);
		\draw [draw, dotted, -, gray] (24) to (51);
		\draw [draw, dotted, -, gray] (24) to (52);
		\draw [draw, dotted, -, gray] (25) to (52);
		\draw [draw, dotted, -, gray] (52) to (79);
		\draw [draw, dotted, -, gray] (49) to (79);
		\draw [draw, dotted, -, gray] (46) to (79);
		\draw [draw, dotted, -, gray] (51) to (78);
		\draw [draw, dotted, -, gray] (48) to (78);
		\draw [draw, dotted, -, gray] (45) to (78);
		\draw [draw, dotted, -, gray] (50) to (77);
		\draw [draw, dotted, -, gray] (47) to (77);
		\draw [draw, dotted, -, gray] (44) to (77);
		\draw [draw, dotted, -, gray] (49) to (76);
		\draw [draw, dotted, -, gray] (46) to (76);
		\draw [draw, dotted, -, gray] (43) to (76);
		\draw [draw, dotted, -, gray] (48) to (75);
		\draw [draw, dotted, -, gray] (45) to (75);
		\draw [draw, dotted, -, gray] (42) to (75);
		\draw [draw, dotted, -, gray] (47) to (74);
		\draw [draw, dotted, -, gray] (44) to (74);
		\draw [draw, dotted, -, gray] (41) to (74);
		\draw [draw, dotted, -, gray] (46) to (73);
		\draw [draw, dotted, -, gray] (43) to (73);
		\draw [draw, dotted, -, gray] (40) to (73);
		\draw [draw, dotted, -, gray] (45) to (72);
		\draw [draw, dotted, -, gray] (42) to (72);
		\draw [draw, dotted, -, gray] (39) to (72);
		\draw [draw, dotted, -, gray] (43) to (70);
		\draw [draw, dotted, -, gray] (40) to (70);
		\draw [draw, dotted, -, gray] (37) to (70);
		\draw [draw, dotted, -, gray] (42) to (69);
		\draw [draw, dotted, -, gray] (39) to (69);
		\draw [draw, dotted, -, gray] (36) to (69);
		\draw [draw, dotted, -, gray] (41) to (68);
		\draw [draw, dotted, -, gray] (38) to (68);
		\draw [draw, dotted, -, gray] (35) to (68);
		\draw [draw, dotted, -, gray] (40) to (67);
		\draw [draw, dotted, -, gray] (37) to (67);
		\draw [draw, dotted, -, gray] (34) to (67);
		\draw [draw, dotted, -, gray] (39) to (66);
		\draw [draw, dotted, -, gray] (36) to (66);
		\draw [draw, dotted, -, gray] (33) to (66);
		\draw [draw, dotted, -, gray] (38) to (65);
		\draw [draw, dotted, -, gray] (35) to (65);
		\draw [draw, dotted, -, gray] (32) to (65);
		\draw [draw, dotted, -, gray] (37) to (64);
		\draw [draw, dotted, -, gray] (36) to (63);
		\draw [draw, dotted, -, gray] (34) to (61);
		\draw [draw, dotted, -, gray] (33) to (60);
		\draw [draw, dotted, -, gray] (32) to (59);
		\draw [draw, dotted, -, gray] (31) to (58);
		\draw [draw, dotted, -, gray] (34) to (64);
		\draw [draw, dotted, -, gray] (33) to (63);
		\draw [draw, dotted, -, gray] (31) to (61);
		\draw [draw, dotted, -, gray] (30) to (60);
		\draw [draw, dotted, -, gray] (29) to (59);
		\draw [draw, dotted, -, gray] (28) to (58);
		\draw [draw, dotted, -, gray] (31) to (64);
		\draw [draw, dotted, -, gray] (30) to (63);
		\draw [draw, dotted, -, gray] (28) to (61);
		\draw [draw, dotted, -, gray] (61) to (106);
		\draw [draw, dotted, -, gray] (60) to (105);
		\draw [draw, dotted, -, gray] (59) to (104);
		\draw [draw, dotted, -, gray] (58) to (103);
		\draw [draw, dotted, -, gray] (70) to (106);
		\draw [draw, dotted, -, gray] (69) to (105);
		\draw [draw, dotted, -, gray] (68) to (104);
		\draw [draw, dotted, -, gray] (67) to (103);
		\draw [draw, dotted, -, gray] (66) to (102);
		\draw [draw, dotted, -, gray] (65) to (101);
		\draw [draw, dotted, -, gray] (64) to (100);
		\draw [draw, dotted, -, gray] (63) to (99);
		\draw [draw, dotted, -, gray] (62) to (98);
		\draw [draw, dotted, -, gray] (61) to (97);
		\draw [draw, dotted, -, gray] (60) to (96);
		\draw [draw, dotted, -, gray] (59) to (95);
		\draw [draw, dotted, -, gray] (58) to (94);
		\draw [draw, dotted, -, gray] (79) to (106);
		\draw [draw, dotted, -, gray] (78) to (105);
		\draw [draw, dotted, -, gray] (77) to (104);
		\draw [draw, dotted, -, gray] (76) to (103);
		\draw [draw, dotted, -, gray] (75) to (102);
		\draw [draw, dotted, -, gray] (74) to (101);
		\draw [draw, dotted, -, gray] (73) to (100);
		\draw [draw, dotted, -, gray] (72) to (99);
		\draw [draw, dotted, -, gray] (71) to (98);
		\draw [draw, dotted, -, gray] (70) to (97);
		\draw [draw, dotted, -, gray] (69) to (96);
		\draw [draw, dotted, -, gray] (68) to (95);
		\draw [draw, dotted, -, gray] (67) to (94);
		\draw [draw, dotted, -, gray] (25) to (108);
		\draw [draw, dotted, -, gray] (26) to (108);
		\draw [draw, dotted, -, gray] (49) to (110);
		\draw [draw, dotted, -, gray] (50) to (111);
		\draw [draw, dotted, -, gray] (51) to (112);
		\draw [draw, dotted, -, gray] (48) to (109);
		\draw [draw, dotted, -, gray] (51) to (109);
		\draw [draw, dotted, -, gray] (52) to (110);
		\draw [draw, dotted, -, gray] (53) to (111);
		\draw [draw, dotted, -, gray] (108) to (112);
		\draw [draw, dotted, -, gray] (108) to (109);
		\draw [draw, dotted, -, gray] (109) to (81);
		\draw [draw, dotted, -, gray] (110) to (82);
		\draw [draw, dotted, -, gray] (111) to (83);
		\draw [draw, dotted, -, gray] (112) to (84);
		\draw [draw, dotted, -, gray] (63) to (81);
		\draw [draw, dotted, -, gray] (64) to (82);
		\draw [draw, dotted, -, gray] (65) to (83);
		\draw [draw, dotted, -, gray] (66) to (84);
		\draw [draw, dotted, -, gray] (67) to (85);
		\draw [draw, dotted, -, gray] (68) to (86);
		\draw [draw, dotted, -, gray] (69) to (87);
		\draw [draw, dotted, -, gray] (70) to (88);
		\draw [draw, dotted, -, gray] (71) to (89);
		\draw [draw, dotted, -, gray] (72) to (90);
		\draw [draw, dotted, -, gray] (73) to (91);
		\draw [draw, dotted, -, gray] (74) to (92);
		\draw [draw, dotted, -, gray] (75) to (93);
		\draw [draw, dotted, -, gray] (72) to (81);
		\draw [draw, dotted, -, gray] (73) to (82);
		\draw [draw, dotted, -, gray] (74) to (83);
		\draw [draw, dotted, -, gray] (75) to (84);
		\draw [draw, dotted, -, gray] (76) to (85);
		\draw [draw, dotted, -, gray] (77) to (86);
		\draw [draw, dotted, -, gray] (78) to (87);
		\draw [draw, dotted, -, gray] (79) to (88);
		\draw [draw, dotted, -, gray] (80) to (89);
		\draw [draw, dotted, -, gray] (109) to (90);
		\draw [draw, dotted, -, gray] (110) to (91);
		\draw [draw, dotted, -, gray] (111) to (92);
		\draw [draw, dotted, -, gray] (112) to (93);
	\end{pgfonlayer}
\end{tikzpicture}
}
\caption{A-TCN architecture (kernel size: 3, dilation factors: 1,3,9).}
\label{fig:A-TCN}
\end{figure*}

We approached the diacritics restoration problem as a character-sequence labeling task.
We chose the output labels as the set of characters in each alphabet.
An alternative way to model the restoration task could have been to produce the possible diacritical marks (including the empty mark) on the output side.
Our choice is motivated by the expectation that the model's scope could be expanded, and it might be able to correct other local errors in the text, not only missing diacritical marks. 

The neural network architecture we considered for sequence labeling are Temporal Convolutional Networks (TCNs).
TCNs are a generic family of models with notable examples including WaveNet \cite{WaveNet}.
TCNs are 1D fully convolutional networks, where the convolutions are causal, and at time $t$ output is produced in each layer by the convolution of input elements from time $t-1$ and earlier \cite{TCN}.

To increase the effective size of the convolutional windows, dilated convolutions can be used \cite{dilated}.
The network is built with dilation factors which increase exponentially by the depth of the network (Fig. \ref{fig:TCN}).
This ensures that the window on the input sequence, which the network can utilize for the inference of a given label, also increases exponentially.

TCNs also contain residual connections \cite{residual}.
A residual block involves a series of transformations, the result of which are then added to the input.
The transformation consists of a dilated convolution followed by a normalization layer, activation function, and dropout.
This is repeated $b$ times (typically $b=2$).

TCNs work well for applications where information flow from the future is not permitted.
For diacritics restoration it is essential to incorporate future context as well as past context.
To achieve this, the base TCN architecture has to be slightly modified as seen in Fig. \ref{fig:A-TCN}.
The modified TCN architecture is called acausal TCN, or A-TCN for short \cite{A-TCN}.

\section{ONNX.js compatibility}

Our model is compatible with ONNX \cite{ONNX}, a cross-platform neural network format.
ONNX serves as an intermediary format, it can be imported by ONNX.js \cite{ONNX.js},  a JavaScript library, which makes it possible to run our model in the browser.
Inference happens on the clients device, making use of the clients graphical processor with the help of WebGL. 

Converting a model to work with ONNX.js requires some care.
For example LSTMs are not supported yet, and even 1D convolutions have to be simulated with 2D convolutions.
Although they are mathematically equivalent, we found that training the model in PyTorch is much more effective if spatial dimension is reduced to 1 in the 2D convolution (instead of reducing the feature size to 1). 

Another difficulty is that the model allows arbitrary input lengths, but in ONNX.js the first inference fixes the input sequence length.
The solution is to dynamically reload the model.
If the input is longer than the current limit, the model is reloaded with double length.

Our demonstration web page with diacritics restoration for four  Central European languages is available at the url\phantom{x}\url{https://web.cs.elte.hu/~csbalint/diacritics/demo.html?lang=en}.

\section{Datasets}

The data for training diacritics restoration can be generated in a self-supervised fashion.
Grammatically correct sentences from the target language provide the annotated data, which means that the removal of the diacritical marks provide the input.

\begin{table*}[ht!]
    \centering
    \begin{tabular}{@{}c!{\vrule width 0.75pt}ccc|ccc@{}}
        \noalign{\hrule height 0.75pt}
        & \multicolumn{3}{c|}{Train} & \multicolumn{3}{c}{Dev}\\
        ~Language ~ & \multicolumn{1}{c}{Sequences} &  Avg.seq.len.~ & \multicolumn{1}{c|}{Characters} & ~Sequences~ &  Avg.seq.len. & Characters\\
        \noalign{\hrule height 0.75pt}
         Cze & \phantom{1}946\,\si{\k}  & $107.6$ & 101.8\,\si{\M} & \SI{14.5}{\k} & $114.4$ & \SI{1.66}{\M}\\
         Hun & 1287\,\si{\k} & $108.3$ & 139.3\,\si{\M} & \SI{14.7}{\k} & $120.7$ & \SI{1.77}{\M}\\
         Pol & 1063\,\si{\k} & $116.2$ & 123.6\,\si{\M} & \SI{14.8}{\k} & $121.3$ & \SI{1.80}{\M}\\
         Svk & \phantom{1}609\,\si{\k}  & $106.7$ & \phantom{1}65.1\,\si{\M} & \SI{14.9}{\k} & $114.7$ & \SI{1.71}{\M}\\
         \noalign{\hrule height 0.75pt}
    \end{tabular}
    \caption{Statistics of the LINDAT datasets.}
    \label{dataLINDAT}
\end{table*}

\begin{table*}[ht!]
    \centering
    \begin{tabular}{c!{\vrule width 0.75pt}cc|cc|cc|cc}
        \noalign{\hrule height 0.75pt}
        &&& \multicolumn{2}{c|}{Unambiguous} & \multicolumn{2}{c|}{Ambiguous} & \multicolumn{2}{c}{Ratio}\\
        Corpus  & Sequences & \multicolumn{1}{c|}{Words} & \multicolumn{1}{c}{Words} & Bases & \multicolumn{1}{c}{Words} & \multicolumn{1}{c|}{Bases} & Words & Bases\\
        \noalign{\hrule height 0.75pt}
        HunWeb1 & \SI{649}{\k}  & \phantom{1}35.7\,\si{\M}  & \phantom{1}18.2\,\si{\M}  & \SI{979}{\k}  & \phantom{1}17.6\,\si{\M}  & \phantom{1}29.3\,\si{\k}  & $1.032$ & $33.5$\\
        HunWeb2 & \SI{6.16}{\M} & 403.0\,\si{\M} & 118.6\,\si{\M} & \SI{4.51}{\M} & 284.4\,\si{\M} & 179.2\,\si{\k} & $0.417$ & $25.2$\\
        \noalign{\hrule height 0.75pt}
    \end{tabular}
    \caption{Word ambiguity statistics of the Webcorpus-based datasets for Hungarian.}
    \label{word_ambiguity}
\end{table*}

We used the datasets provided by \newcite{naplava2018} for training on four Central European languages (Czech, Hungarian, Polish and Slovak).
We will refer to these datasets as LINDAT. 
The datasets were cleaned up by removing the sentences containing exotic characters (we considered the character exotic if applying the unidecode Python function on the character yielded a string more than one character long). 
We also cut off all the sentences to a maximum length of 500.
Table \ref{dataLINDAT} shows the statistics of the datasets.

Two additional corpora were considered for training and evaluating on Hungarian. 
A model was trained on the dataset built from Hungarian Webcorpus 2.0, hereinafter referred to as HunWeb2 \cite{NemeskeyPhD2020}.
The models were also evaluated on the dataset built from the earlier Hungarian Webcorpus (HunWeb1) by \newcite{halacsy2004creating}.  

Each corpus contains a large collection of Hungarian text documents. 
To prepare the data, we extracted sentences from each document until we reached a length limit of 500.
After extracting the sequences, we random sampled them, and created the train-dev cuts (Table \ref{dataHunWeb12}).
We also cleaned up the data by removing all sequences not containing enough diacritical marks, as some part of the corpus contains sentences wich partially or completely lack the proper diacritical marks.
In the case of HunWeb2, we used the "2017-2018" part of the Common Crawl subcorpus.
We decided not to split up the documents to multiple chunks or sentences, as we have found that there can be a lot of repeated sentences within one document, which would skew the evaluation metrics.

\begin{table}[ht!]
    \centering
    \begin{tabular}{cc!{\vrule width 0.75pt}ccc}
        \noalign{\hrule height 0.75pt}
        \multicolumn{2}{c!{\vrule width 0.75pt}}{\multirow{2}{*}{Dataset}}& \multirow{2}{*}{Seqs} & Avg.            & \multirow{2}{*}{Chars} \\
        \multicolumn{2}{c!{\vrule width 0.75pt}}{}                        &                       & seq. len.       &     \\
        \noalign{\hrule height 0.75pt}
        HunWeb1                  & Dev   & \SI{10}{\k}   & $409.3$         & \SI{4.09}{\M} \\\hline
        \multirow{2}{*}{HunWeb2} & Train & \SI{6.16}{\M} & $474.0$         & \SI{2.92}{\G}  \\
                                 & Dev   & \SI{10}{\k}   & $474.1$         & \SI{4.74}{\M} \\
        \noalign{\hrule height 0.75pt}
    \end{tabular}
    \caption{Statistics of the additional datasets for Hungarian.}
    \label{dataHunWeb12}
\end{table}

\section{Word Ambiguity}

Ambiguous words pose a limit to dictionary-based solutions for solving diacritics restoration.
For example we have several options when we want to diacritize the Hungarian word \emph{koros}:
\emph{körös} (containing circles or a Hungarian river), \emph{kóros} (sick), \emph{kórós} (containing weeds), \emph{koros} (old) and even \emph{kőrös} (geographical name).

We analyzed the Hungarian Webcorpus-based datasets in terms of apparent word ambiguity.
Let us call the base of a word the word we get after removing the diacritical marks from it.
We categorized a base unambiguous if the data contained only one diacritized version of it.
Similarly, a word was categorized ambiguous if multiple diacritized forms existed in the data.
The ambiguity of a word may be due to grammar or to an error in the corpus, even after the cleanup step was performed to decrease the number of such false positives.
Unambiguous words can be diacritized with a dictionary-based approach.

In Table \ref{word_ambiguity} we see the statistics related to ambiguous and unambiguous words in the datasets.
There are similar amounts of ambiguous and unambiguous words in the data 
(though as the number of sequences increases, the chance for false positives and rare variants also increases), but the ambiguous words come from a much smaller set of bases.

The metric of ambiguous word accuracy is dependent on which words are classified as ambiguous, which makes it unsuitable to compare the performance of different models.
Nevertheless when compared to (alpha) word accuracy, we saw that ambiguous word accuracy was higher at the beginning of the training, but as the model improved the two metrics switched places.
This might be explained by the numbers in Table \ref{word_ambiguity}, as the ambiguous bases are harder to correctly diacritize on the long run, but there are substantially more unambiguous bases, and the model might need more examples to memorize them.

\section{Experimental Setup}

\begin{table*}[ht]
\begin{center}
    \begin{tabular}{ccc!{\vrule width 0.75pt}cccc}
        \noalign{\hrule height 0.75pt}
        Model & Train data & ~ Eval data ~ & ~Character~ &  Vowel~ & Alpha-word~ & Sequence \\
        \noalign{\hrule height 0.75pt}
        \multirow{3}{*}{Copy}      & \multirow{3}{*}{}        & HunWeb1 & $0.8979$        & $0.6929$        & $0.4768$        & $0.0000$ \\ 
                                   &                          & HunWeb2 & $0.9020$        & $0.7042$        & $0.4997$        & $0.0000$ \\
                                   &                          & LINDAT  & $0.9043$        & $0.7134$        & $0.5093$        & $0.0269$ \\\hline
        \multirow{3}{*}{Hunaccent} & \multirow{3}{*}{HunWeb1} & HunWeb1 & $0.9886$        & $0.9657$        & $0.9207$        & $0.0398$ \\ 
                                   &                          & HunWeb2 & $0.9855$        & $0.9563$        & $0.9049$        & $0.0087$ \\
                                   &                          & LINDAT  & $0.9834$        & $0.9509$        & $0.8934$        & $0.2732$ \\\hline
        \multirow{3}{*}{Dictionary}& \multirow{3}{*}{HunWeb2} & HunWeb1 & $0.9960$        & $0.9879$        & $0.9772$        & $0.3511$ \\ 
                                   &                          & HunWeb2 & $0.9965$        & $0.9894$        & $0.9791$        & $0.3329$ \\
                                   &                          & LINDAT  & $0.9942$        & $0.9831$        & $0.9698$        & $0.6551$ \\\hline
        \multirow{3}{*}{A-TCN}     & \multirow{3}{*}{HunWeb2} & HunWeb1 & \textbf{0.9987} & \textbf{0.9961} & \textbf{0.9907} & \textbf{0.6574} \\
                                   &                          & HunWeb2 & \textbf{0.9988} & \textbf{0.9964} & \textbf{0.9916} & \textbf{0.6424} \\
                                   &                          & LINDAT  & $0.9974$        & \textbf{0.9941} & \textbf{0.9862} & \textbf{0.8087} \\\hline
        \multirow{3}{*}{A-TCN}     & \multirow{3}{*}{LINDAT}  & HunWeb1 & $0.9950$        & $0.9850$ & $0.9649$ & $0.2683$ \\
                                   &                          & HunWeb2 & $0.9945$        & $0.9834$ & $0.9621$ & $0.1556$ \\
                                   &                          & LINDAT  & \textbf{0.9975} & $0.9925$ & $0.9824$ & $0.7890$ \\
        \noalign{\hrule height 0.75pt}\\
    \end{tabular}
    \caption{Accuracy comparison for Hungarian diacritics restoration between the baseline (Hunaccent) and our model (A-TCN). We used the pretrained Hunaccent  model provided by the authors. The numbers indicate the results on non-augmented, fully dediacritized input.}
    \label{results_big}
\end{center}
\end{table*}

In terms of model architecture we used the following hyperparameters.
The character embedding dimension was set to 50.
After the embedding, the vectors are upsampled to dimension 250, which is the channel size.
Upsampling is done by an $1\times 1$ convolution after permuting the dimensions from $50\times 1\times n$ to $1\times 50\times n$, where $n$ is the sequence length. This effectively means that the input is concatenated by itself $5$ times over, allowing for a scalar multiplier for each copy. 
This simpler approach was chosen instead of the usual upsampling for performance reasons.

The network contains 4 residual block layers with dilation factors of 1,2,4, and 8, respectively.
Each block contains 2 convolutional layers, each followed by batch normalization, ReLU, and spatial dropout layers with a rate of 0.2, respectively.
The convolutions have a kernel size of 5.
In the convolutions, zero padding is used to ensure that the output is the same length as the input.

We augmented the training data before each epoch in the training.
If a character had a diacritical mark, we removed it with a probability of 80\%.
In real world use, the absence of diacritical marks might only be partial.

When trained on HunWeb2 we limited each epoch to 100000 sequences.
The batches contain the same 200 sequences every time (augmentation is applied each time a batch is accessed), but in each epoch we train on a random 500 batches.

The model implemented in PyTorch was trained on 4 Nvidia RTX 2080 Ti graphics cards.
Training took approximately one day per model.
Our model is available at
{\small\url{https://github.com/aielte-research/Diacritics_restoration}}.

\section{Results}

We calibrated our model lightweight enough to be converted to HTML.
For Hungarian we took Hunaccent \cite{hunaccent} as a direct comparison.
Hunaccent is decision tree based, and it shares our goal to implement a small footprint restorator.
Moreover, it also can be run locally in a browser.
We considered the pretrained Hunaccent model provided by the authors \newcite{hunaccent}.
To ensure a fair comparison, we set up our model to have a size similar to the 12.1 MB of the trained model of Hunaccent.
The raw ONNX file of our trained model is 10.11 MB and our demo HTML file is 13.49 MB.
The HTML file contains the ONNX file as a Base64 encoded string.

Compared to Hunaccent, our model achieved significantly better results in all of the metrics we considered.
We also measured the performance of a simple dictionary based method. 
From the HunWeb2-based training data, we created a word dictionary containing each word base we encountered.
The most frequent diacritization was chosen for each base.
\newcite{naplava2018} reports an alpha word accuracy of $0.9902$ on Hungarian (LINDAT).
Their model is LSTM-based and has a reported size of around 30 MB. 

Table \ref{results_big} contains the results.
We also added a line called \emph{Copy}, which measures the accuracy what we get if we simply copy the input without adding any diacritical marks.
\textit{Character} accuracy measures the ratio of the correct characters in the output.
\textit{Important character} accuracy is measured on characters for which  diacritical marks are applicable.
In the case of the Hungarian language, these characters are the vowels.
\textit{Alpha-word} accuracy is measured by the ratio of the correct words in the output, where only the words are considered which contain at least one alphabetical character.
\textit{Sequence} accuracy is measured by the ratio of flawless sequences, which is inversely proportional to the average length of the sequences.

In Table \ref{results_augment} we can see the effect of the augmentation.
Hunaccent performs better on data where all of the diacritics are missing, while our model performs slightly better, but almost the same when we remove only about 80\% of the diacritical marks.

\begin{table}[H]
    \centering
    \begin{tabular}{cc!{\vrule width 0.75pt}cc}
        \noalign{\hrule height 0.75pt}
        Model                         &  Eval. task & ~Vowel~         & Alpha-word\\
        \noalign{\hrule height 0.75pt}
        \multirow{2}{*}{Hunaccent}    & Aug.         & $0.9441$        & $0.8785$\\
                                      & Non-aug.     & \textbf{0.9563} & \textbf{0.9048} \\\hline
        \multirow{2}{*}{A-TCN}        & Aug.         & \textbf{0.9967} & \textbf{0.9925} \\
                                      & Non-aug.     & $0.9964$        & $0.9915$ \\
        \noalign{\hrule height 0.75pt}
    \end{tabular}
    \caption{Performance comparison of Hunaccent (normal training) and A-TCN (augmented training) on the augmented and the non-augmented task (HunWeb2). }
    \label{results_augment}
\end{table}

For Hungarian we compared the datasets in terms of performance of the trained models (Table \ref{results_big}).
Our tests indicate that our HunWeb2-based dataset yields better results.
This is partly due to the size difference between the training data.
When trained on a smaller size HunWeb2-based dataset, the model still performed better.
This might be explained by Table \ref{results_overfit}, as the model seems to overfit when trained on LINDAT.
The train and dev data are likely not independent enough.

\begin{table}[H]
    \centering
    \begin{tabular}{cc!{\vrule width 0.75pt}cc}
        \noalign{\hrule height 0.75pt}
        \multicolumn{2}{c!{\vrule width 0.75pt}}{Dataset}& ~Vowel~ & Alpha-word~  \\
        \noalign{\hrule height 0.75pt}
        \multirow{2}{*}{HunWeb2} & Train & $0.9924$ & $0.9828$ \\
                                 & Dev   & $0.9893$ & $0.9764$ \\\hline
        \multirow{2}{*}{LINDAT}  & Train & $0.9922$ & $0.9816$ \\
                                 & Dev   & $0.9925$ & $0.9824$ \\
        \noalign{\hrule height 0.75pt}
    \end{tabular}
    \caption{Train and dev accuracies of the same model trained on HunWeb2 and LINDAT. The model seems to overfit on LINDAT.}
    \label{results_overfit}
\end{table}

The performance of our model on four Central European languages from the LINDAT corpus can be seen in Table \ref{resultsv4}.
The results indicate that our model is language-agnostic and works well for its size for multiple different languages.  
The alpha-word accuracies are slightly below the ones reported by \cite{naplava2018}.

\begin{table}[H]
    \centering
    \begin{tabular}{c!{\vrule width 0.75pt}cccc}
        \noalign{\hrule height 0.75pt}
        Lang. & Chr. &  Imp. Chr. & $\alpha$-word & Seq. \\
        \noalign{\hrule height 0.75pt}
        Cze & $0.9966$ & $0.9944$ & $0.9783$ & $0.7344$ \\
        Hun & $0.9975$ & $0.9925$ & $0.9824$ & $0.7890$ \\
        Pol & $0.9987$ & $0.9970$ & $0.9903$ & $0.8810$ \\
        Svk & $0.9966$ & $0.9947$ & $0.9784$ & $0.7420$ \\\noalign{\hrule height 0.75pt}
    \end{tabular}
    \caption{Accuracies on languages trained on the LINDAT dataset.}
    \label{resultsv4}
\end{table}

\section{Error Analysis}

\newcolumntype{I}{!{\vrule width 0.75pt}}
\newlength\savedwidth
\newcommand\whline[1]{\noalign{\global\savedwidth\arrayrulewidth\global\arrayrulewidth 0.75pt}%
\cline{#1}
\noalign{\global\arrayrulewidth\savedwidth}}

\newcommand\ExtraSep
{\dimexpr\cmidrulewidth+\aboverulesep+\belowrulesep\relax}

\begin{table*}[ht!]
    \centering
    \begin{tabular}{cIcIc|c|c|c|c|c|c|c|c|c|c|c|c|cI}
        \multicolumn{1}{c}{}&\multicolumn{15}{c}{Predicted Vowel}\\
        \multicolumn{15}{c}{\vspace{-3mm}}\\
        \whline{2-8}  \whline{10-16}
        \multirow{11}{4mm}{\rotatebox[origin=c]{90}{Actual Vowel}} &
                                 & o             &                     ó              & ö             &   \multicolumn{2}{c|}{ő}   & \multicolumn{1}{cI}{\scriptsize{TPR}}&
        \multicolumn{1}{cI}{}&   & u             & \multicolumn{2}{c|}{ú}             & ü             &     ű                      & \scriptsize{TPR}
        \\ \whline{2-8} \whline{10-16}
                             & o & \SI{156}{\k}  & $302$                              & $164$         & \multicolumn{2}{c|}{$118$} & \multicolumn{1}{cI}{$0.996$}&
        \multicolumn{1}{cI}{}& u & \SI{43.2}{\k} & \multicolumn{2}{c|}{$91$}          & $54$          & $23$                       & $0.996$                     
        \\\cline{2-8}\cline{10-16}
                             & ó & $297$         & \SI{42.8}{\k}                      & $28$          & \multicolumn{2}{c|}{$77$}  & \multicolumn{1}{cI}{$0.991$}&
        \multicolumn{1}{cI}{}& ú & $82$          & \multicolumn{2}{c|}{\SI{12.4}{\k}} & $3$           & $14$                       & $0.992$
        \\\cline{2-8}\cline{10-16}
                             & ö & $166$         & $28$                               & \SI{42.6}{\k} & \multicolumn{2}{c|}{$92$}  & \multicolumn{1}{cI}{$0.993$} &
        \multicolumn{1}{cI}{}& ü & $72$          & \multicolumn{2}{c|}{$21$}          & \SI{23.6}{\k} & $46$                       & $0.994$
        \\\cline{2-8}\cline{10-16}
                             & ő & $67$          & $68$                               & $46$  & \multicolumn{2}{c|}{\SI{38.3}{\k}} & \multicolumn{1}{cI}{$0.995$}&
        \multicolumn{1}{cI}{}& ű & $19$          & \multicolumn{2}{c|}{$15$}          & $24$          & $8263$                     & $0.993$
        \\\cline{2-7}\whline{8-8}\cline{10-15}\whline{16-16}
              & \scriptsize{PPV} & $0.997$       & $0.991$                            & $0.994$       &\multicolumn{2}{cI}{$0.993$}&\multicolumn{1}{c}{}         &
        \multicolumn{1}{cI}{}&\scriptsize{PPV}& $0.996$& \multicolumn{2}{c|}{$0.990$} & $0.997$       &\multicolumn{1}{cI}{$0.990$}&\multicolumn{1}{c}{}
        \\\whline{2-7}  \whline{10-15}
        \multicolumn{15}{c}{\vspace{-2mm}}
        \\\whline{2-5}  \whline{7-11} \whline{13-16}
                             &                              & a            & á                                 & \multicolumn{1}{cI}{\scriptsize{TPR}}
                             &    \multicolumn{1}{cI}{}&    & e            & \multicolumn{2}{c|}{é}            & \multicolumn{1}{cI}{\scriptsize{TPR}}  &
        \multicolumn{1}{cI}{}&                              & i            & í                                 & \scriptsize{TPR}
        \\\whline{2-5}  \whline{7-11} \whline{13-16}
                             & a                            & \SI{345}{\k} & $856$                             & \multicolumn{1}{cI}{$0.998$}           &
        \multicolumn{1}{cI}{}& e                            & \SI{391}{\k} & \multicolumn{2}{c|}{$926$}        & \multicolumn{1}{cI}{$0.998$}           &
        \multicolumn{1}{cI}{}& i                            & \SI{169}{\k} &  $220$                            & $0.999$
        \\\cline{2-5}\cline{7-11}\cline{13-16}
                             & á                            & $826$        & \SI{138}{\k}                      & \multicolumn{1}{cI}{$0.994$}           &
        \multicolumn{1}{cI}{}& é                            & $1313$       & \multicolumn{2}{c|}{\SI{132}{\k}} & \multicolumn{1}{cI}{$0.990$}           &
        \multicolumn{1}{cI}{}& í                            & $159$        & \SI{25.0}{\k}                     & $0.994$
        \\\cline{2-4}\whline{5-5}\cline{7-10}\whline{11-11}\cline{13-15} \whline{16-16}
                             & \scriptsize{PPV}             & $0.998$      & \multicolumn{1}{cI}{$0.994$}      & \multicolumn{1}{c}{}                   &
        \multicolumn{1}{cI}{}& \scriptsize{PPV}             & $0.997$      & \multicolumn{2}{cI}{$0.993$}      & \multicolumn{1}{c}{}                   &
        \multicolumn{1}{cI}{}& \scriptsize{PPV}             & $0.999$      & \multicolumn{1}{cI}{$0.991$}
        \\\whline{2-4}  \whline{7-10} \whline{13-15}
    \end{tabular}
 
    \caption{Vowel confusion matrix}
    \label{vowel_conf}
\end{table*}

The confusion matrix of the A-TCN model (trained and evaluated on HunWeb2) can be seen in Table \ref{vowel_conf}.
Even though our model can output every character in the vocabulary at each position, the only confused characters were vowels with the same base.
We included precision (PPV) an recall (TPR) in the table.
The overall weighted F1 score for vowels is $0.996$. 

We performed a small-scale manual evaluation of the A-TCN model.
After inference on the evaluation dataset, we selected 500 random errors to be manually classified in the following categories.

\begin{enumerate}[itemsep=-3pt, topsep=0pt, partopsep=0pt]
    \item The error happens due to a corpus error.
    \item The error is false positive due to a corpus error, the model output is the correct form.
    \item The input is ambiguous at word level, but the model output does not fit grammatically in the sentence.
    \item The output is not wrong grammatically, but does not agree with the wider context of the text.
    \item Though the model output and the ground truth are different, they both are adequate.
    \item The error occurred in a named entity.
    \item None of the above.
\end{enumerate}

\begin{table}[H]
    \centering
    \begin{tabular}{c!{\vrule width 0.75pt}c}
        \noalign{\hrule height 0.75pt}
        Error class & Ratio \\
        \noalign{\hrule height 0.75pt}
        1. Corpus error & $0.062$ \\
        2. Corrected corpus error & $0.128$ \\
        3. Word Ambiguous Input & $0.186$ \\
        4. Grammar Ambiguous Input & $0.158$ \\
        5. Context Ambiguous Input & $0.124$ \\ 
        6. Named Entity & $0.256$ \\
        7. Incorrect Output & $0.126$ \\\noalign{\hrule height 0.75pt}
    \end{tabular}
    
    \caption{Error classes of the Hungarian A-TCN model.}
    \label{err_cls}
\end{table}

According to the manual evaluation (Table \ref{err_cls}) around 30\% of the errors belong to categories 3 and 7.
We can reasonably expect to reduce these errors by increasing the size of the model, both to increase the perceived vocabulary of the model, and also to enable a larger context window to draw information from, as some of the grammatical context is likely too far away for the model with the current hyperparameters.
Named entity errors are a bit harder to reduce, since they are often less frequent or more ambiguous in the corpus.
Errors due to ambiguous input in terms of grammar could be harder to reduce as they sometimes require more insight. 

\section{Conclusion}

We presented a neural network model of small size based on 1D convolutions for diacritics restoration.
Furthermore, the model is ONNX.js compatible, so it can even be used in a web browser.
The model was evaluated on four Central European languages and it performed similarly well compared to other larger models and outperformed models of similar size. In the case of the Hungarian language, we considered three data sets and studied the generalizing power of the model between data sets.

Further research is needed to expand the applicability of the model to correcting general errors in texts, including spelling.
We plan to train a larger, but still browser-compatible model, and plan to further improve the model architecture.

\section{Acknowledgments}

The research was partially supported by the Ministry of Innovation and Technology NRDI Office within the framework of the Artificial Intelligence National Laboratory Program, the Hungarian National Excellence Grant 2018-1.2.1-NKP-00008 and the Thematic Excellence Programme TKP2021-NKTA-62.

The second author was supported by project "Application Domain Specific Highly Reliable IT Solutions" implemented with the support provided from the National Research, Development and Innovation Fund of Hungary, financed under the Thematic Excellence Programme TKP2020-NKA-06 (National Challenges Subprogramme) funding scheme.

We would like to thank Dániel Varga for drawing our attention to the problem of lightweight diacritics reconstruction, and Judit Ács for her help with NLP issues.

\section{Bibliographical References}\label{reference}
\label{main:ref}
\bibliographystyle{lrec2022-bib}
\bibliography{ms}

\section{Language Resource References}
\label{lr:ref}
\bibliographystylelanguageresource{lrec2022-bib}
\bibliographylanguageresource{languageresource}

\color{white}
\citelanguageresource{HunWeb1}
\citelanguageresource{HunWeb2}
\citelanguageresource{LINDAT}

\end{document}